\documentclass[journal]{IEEEtran}

\ifCLASSINFOpdf
\else
\fi

\usepackage[pagebackref=true,breaklinks=true,colorlinks,bookmarks=false]{hyperref}

\usepackage[pdftex]{graphicx}
\usepackage{bm}
\usepackage{amsmath}
\usepackage{amssymb}
\usepackage{algorithm}
\usepackage{mathrsfs}
\usepackage{enumitem} 
\usepackage{multirow}
\usepackage{tabularx}
\usepackage{adjustbox}
\usepackage{makecell}
\usepackage{color}
\usepackage{xcolor}
\usepackage{verbatim}

\usepackage{algorithm}
\usepackage{algpseudocode}
\usepackage{graphics}
\usepackage{epsfig}
\usepackage[mathscr]{euscript}
\usepackage{siunitx}
\usepackage{caption}
\usepackage{url}

\graphicspath{image}

\def\bX{{\bf X}}

\begin{document}
\title{Unsupervised Person Re-identification with Stochastic Training Strategy}

\author{Tianyang Liu, Yutian Lin and Bo Du
\thanks{
(Corresponding Author: Yutian Lin and Bo Du)

T. Liu, Y. Lin, and B. Du are with the National Engineering Research Center for Multimedia Software, School of Computer Science, Institute of Artificial Intelligence, and Hubei Key Laboratory of Multimedia and Network Communication Engineering, Wuhan University. (E-mail: 2015301500110@whu.edu.cn; yutian.lin@whu.edu.au; bodu@whu.edu.cn).
}
}

\markboth{Submitted to IEEE TRANSACTIONS ON IMAGE PROCESSING}%
{Shell \MakeLowercase{\textit{et al.}}: Bare Demo of IEEEtran.cls for IEEE Journals}
\maketitle

\begin{abstract} 
Unsupervised person re-identification (re-ID) has attracted increasing research interests because of its scalability and possibility for real-world applications. 
State-of-the-art unsupervised re-ID methods usually follow a clustering-based strategy, which generates pseudo labels by clustering and maintains a memory to store instance features and represent the centroid of the clusters for contrastive learning. This approach suffers two problems. First, the centroid generated by unsupervised learning may not be a perfect prototype. Forcing images to get closer to the centroid emphasizes the result of clustering, which could accumulate clustering errors during iterations. Second, previous instance memory based methods utilize features updated at different training iterations to represent one centroid, these features are inconsistent due to the change of encoder.

To this end, we propose an unsupervised re-ID approach with a stochastic learning strategy. Specifically, we adopt a stochastic updated memory, where a random instance from a cluster is used to update the cluster-level memory for contrastive learning. In this way, the relationship between randomly selected pair of images are learned to avoid the training bias caused by unreliable pseudo labels. By picking a sole last seen sample to directly update each cluster center, the stochastic memory is also always up-to-date for classifying to keep the consistency. Besides, to relieve the issue of camera variance, a unified distance matrix is proposed during clustering, where the distance bias from different camera domains is reduced and the variances of identities are emphasized. Our proposed method outperforms the state-of-the-arts in all the common unsupervised and UDA re-ID tasks. The code will be available at \url{https://github.com/lithium770/Unsupervised-Person-re-ID-with-Stochastic-Training-Strategy}.
\end{abstract}


\IEEEpeerreviewmaketitle

\section{Introduction}
   \begin{figure}[ht]
   \centering
    \includegraphics[width=\linewidth]{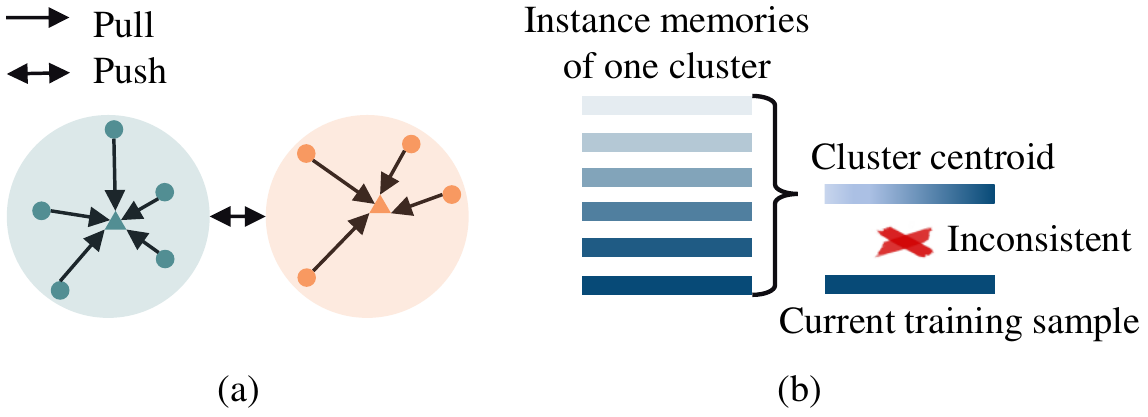}
   \caption{(a) Illustration of the typical clustering-based contrastive re-ID loss. Circles in the same color denote images of the same cluster. Triangles denote the cluster centroid. For each training sample, the model learns to push the instance closer to its own cluster centroid while away from the other centroids. (b) The illustration of the cluster centroid. Rectangles with different shades of color represent instance features of the same cluster that have been stored at different training stages. The darker the color, the newer the model used to extract the features.}
   \label{fig:motivation}
\end{figure}

 \IEEEPARstart{P}{erson} re-identification (re-ID) aims to match identities across disjoint cameras. Due to the need for public safety and intelligent surveillance, it has drawn wide attention in the computer vision community \cite{li2014deepreid,liao2015person,zheng2015scalable}. With the advancement of deep learning, fully-supervised person re-ID has gained great progress these years \cite{varior2016gated, sun2019dissecting,miao2019pose, sun2018beyond, chen2019abd, 7918589, 9259257}. However, these methods rely on a large scale of annotated data, which is not realistic in practical applications. To overcome the scalability issue, recent studies focus on unsupervised settings to learn a re-ID model without annotations. 
The kind of unsupervised methods can be further divided into unsupervised domain adaptation (UDA) \cite{wei2018person,deng2018image,chen2019instance, fu2019self, sun2021unsupervised, 9442325,2020Learning, dai2021dual,zheng2021group,dai2021idm} and fully-unsupervised \cite{lin2019bottom, lin2020unsupervised, zeng2020hierarchical, Chen_2021_ICCV}. This paper addresses fully unsupervised person re-ID, which does not require labeled source datasets and is thus more challenging.

The state-of-the-art unsupervised methods on person re-ID usually follow a clustering-based strategy, which alternates between generating pseudo labels by clustering the unlabeled training data and training the network with the pseudo labels \cite{lin2019bottom,ge2020selfpaced,dai2021dual,zheng2021group,Chen_2021_ICCV}. To conduct contrastive learning \cite{wu2018unsupervised,zhuang2019local,he2020momentum,zheng2021group,Chen_2021_ICCV}, a momentum-updated memory is usually adopted. As in a common pipeline \cite{ge2020selfpaced}, an instance-level memory is usually adopted, where the average instance feature from each cluster is used to represent the centroid of the cluster and is further used for contrastive learning. However, this strategy may cause three problems. First, as shown in Fig. \ref{fig:motivation} (a), training with cluster centroids letting all instances get closer to their centroid, making the clustering result more stable, which will provide less supervision signal for training in the next iteration and accumulate errors in the pseudo labels during iterations. Besides, contrastive learning benefits from observing sufficient negative samples~\cite{jaiswal2021survey}. Replace a large number of negatives with only the cluster centroids would damage the diversity of the contrastive pairs. Second, as shown in Fig. \ref{fig:motivation} (b), an instance feature in memory was updated when it was last seen. If we construct a huge memory for all instances, the inconsistency among representations will be severe due to the change of encoder.

To address the above problems, in this paper, we propose an unsupervised person re-identification method with a stochastic learning strategy. The main idea is, instead of using the cluster centroid as the positive and negative samples, all the instances are taken as candidate pools to provide positive and negative samples randomly. Specifically, our framework adopts a stochastic updated memory for contrastive learning. We initialize cluster centers with a random instance belonging to each cluster, and directly update cluster centers along with training. Then for each training data, the feature embedding is learned to be close/away to last seen instances, instead of the centroid of the whole class. On the one hand, without converging toward a definite target, the accumulated errors caused by noisy pseudo-labels are eliminated, and a large amount of negatives introduces diversity into the model learning. On the other hand, each cluster center is represented with the last seen instances.  The stochastically updated memory makes a more consistent centroid embedding to learn the relation with the current training sample. 

As a minor contribution, we improve the procedure of clustering to generate more reliable pseudo labels. Since the style of images is various across different cameras, we propose a unified distance matrix to relieve the camera variance. 
Specifically, the mean value of target instances' cosine similarities is calculated for each intra-camera and inter-camera case, which implicitly reflects the distance caused by camera variances. At each epoch before clustering, we subtract the corresponding value from the similarities of instance pairs according to their camera labels and then softmax the distance vector of each instance. In this way, the camera variance is reduced and images of the same identity cross cameras will be more likely to be clustered. 
Moreover, we adopt a temporal ensembling based embedding for clustering. Instead of using the feature extracted from the current model, the embedding is further assembled with past temporally information. Specifically, an instance-level memory is adopted during training to save feature embeddings with history information, leading to more robust features.

Our contribution can be summarized in three-fold:

(1) We explore the elements affecting contrastive learning for unsupervised person re-identification, \textit{e.g.} the contrast target, and the consistency between the contrasting samples. A stochastic learning strategy is proposed to learn the relationship between stochastic instances of each cluster and use the up-to-date features to keep the consistency. The effectiveness of the proposed method is quantitatively analyzed.

(2) Camera and temporal information is adopted for more precise clustering. We emphasize the importance of camera variances in unsupervised re-ID, and propose a unified distance matrix to highlight the identity variance while reducing the camera variance. Besides, temporal ensembling is also adopted to build more robust features for clustering.

(3) Extensive experiments are conducted on four common large-scale datasets. Our method significantly outperforms the state-of-the-art unsupervised and UDA re-ID approaches on both person and vehicle datasets.

\section{Related Work}
\textbf{Fully unsupervised person re-ID}. Existing fully unsupervised person re-ID methods usually explore pseudo labels for learning \cite{lin2019bottom, lin2020unsupervised,zeng2020hierarchical, Wang2021camawareproxies}. 
In BUC~\cite{lin2019bottom}, a bottom-up clustering method is proposed to gradually obtain more reasonable clusters for network training. In SSL~\cite{lin2020unsupervised}, the K-nearest neighbours for each training sample are explored to assign soft pseudo labels for training.  HCT~\cite{zeng2020hierarchical} uses hierarchical clustering to generate pseudo labels and sample a part of them for hard-batch triplet loss. These clustering-based methods usually work in a camera-agnostic way, which may result in noisy pseudo labels caused by camera variances. To overcome it, DAL~\cite{chen2018deep} and UGA~\cite{wu2019unsupervised} divide the training task into intra-camera and inter-camera stages. Intra-camera labels have less noise, and knowledge from intra-camera labels can help association across cameras. However, utilizing camera labels to generate more precise pseudo labels has been undervalued. 

Recently, combination with contrastive loss and momentum-updated memory has gained great attention~\cite{ge2020selfpaced, Wang2021camawareproxies}. SPCL~\cite{ge2020selfpaced} is a UDA-based method that can work in both unsupervised domain adaptation and fully unsupervised settings. It adopts an instance-level memory for instance features and contrastive loss. In CAP~\cite{Wang2021camawareproxies}, camera variance is emphasized, where a proxy-level memory is constructed by considering the camera label. This method works well, especially on the dataset with multiple cameras. Based on CAP~\cite{Wang2021camawareproxies}, ICE~\cite{Chen_2021_ICCV} adds a momentum-updated encoder and proposes contrastive learning between input features and momentum representations. Compared with the above methods, our work follows the same clustering-based pipeline. However, we focus more on the contrastive learning strategy, where a stochastic learning strategy is adopted to avoid the accumulation of clustering error, ensure the diversity of negative samples and keep the comparison samples consistent. 

\begin{figure*}[ht]
   \centering
   \includegraphics[width=\linewidth
   ]{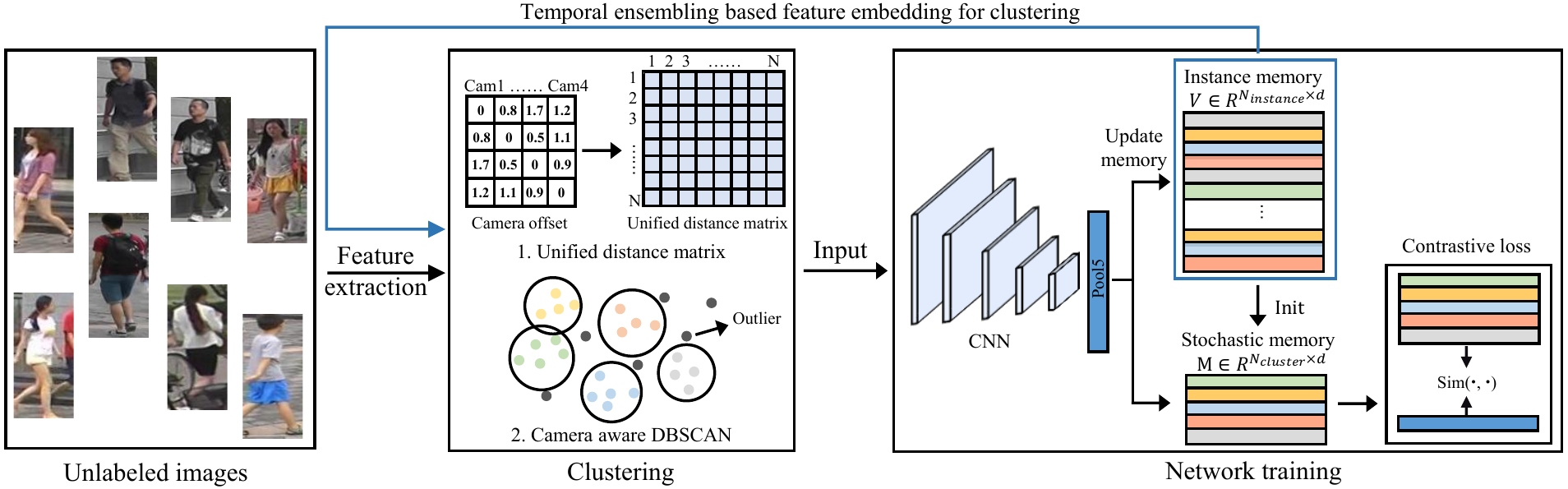}
   \caption{An overview of the proposed method. It iterates between clustering and network training. For initialization, the features of all training samples are extracted to contruct the instance memory for clustering. During training, the stochastically-updated memory is used as a classifier with contrastive loss. The stochastic memory and the instance-level memory are updated along with training. During clustering, features in the instance memory are used to compute the camera offset matrix (here we assume there are 4 cameras) and the unified distance matrix, where the distance is normalized with camera labels. Then the new distance matrix is adopted for DBSCAN to generate pseudo labels. The stochastically updated memory is then re-initialized via pseudo labels and instance memory features.}
   \label{fig:overview}
\end{figure*}

\textbf{Unsupervised domain adaptation (UDA) person re-ID}. UDA-based methods require some labeled source domain datasets and aim to train a model that generalizes well on the unlabeled target domain. They can be divided into two categories: domain translation-based methods~\cite{wei2018person,deng2018image,chen2019instance} and pseudo-label-based methods~\cite{fu2019self,ge2020mutual,zhong2019invariance,ge2020selfpaced,2020Learning,zheng2021group}. The former focuses on transferring knowledge from labeled source domain to unlabeled target domain via performing image-to-image translations. While these methods usually adopt translated instances for training and leave the target data unexploited. The pseudo-label-based methods were found more effective to capture the target distributions. These methods use a unified framework to learn from labeled source datasets and unlabeled target datasets, which repeatedly generate pseudo labels and train the model with pseudo labels. The pseudo labels can be generated by clustering instance features or comparing with exemplar features. The core of these methods is to ensure the reliability of pseudo labels. Among them, SSG~\cite{fu2019self} exploits local features to generate multi-scale pseudo labels. MMT~\cite{ge2020mutual} proposes to improve the robustness of pseudo labels via mutual learning. ECN~\cite{zhong2019invariance} combines exemplar-invariance, camera-invariance and neighbor-invariance in the target domain. SPCL~\cite{ge2020selfpaced} designs cluster reliability criteria to measure the independence and compactness of clusters. Dual-Refinement~\cite{dai2021dual} adopts the offline pseudo label refinement to assign more accurate labels and the online feature refinement to alleviate the effects of noisy supervision signals. GLT~\cite{zheng2021group} combines the label refining algorithm and the group-aware strategy to better correct the noisy pseudo label in an online fashion. In IDM~\cite{dai2021idm}, intermediate domains’ representations are generated by mixing the source and target domains’ hidden representations to narrow the gap between the two domains.

\textbf{Camera variances of person re-ID}. Person images taken by different cameras have various poses, illumination, resolution, etc. In other words, they are subject to different distributions. Different camera labels can cause severe variances between two instances, even they belong to the same identity. Camera variances is an important factor in person re-ID tasks, especially for UDA-based or fully unsupervised setting, in which the pseudo labels will be influenced by camera variances. There are some methods that take camera variances into consideration: CBN~\cite{zhuang2020rethinking} aligns the data in feature space. Camera-aware neighbor invariance learning~\cite{2020Generalizing} divides the neighbor search procedure into the intra-camera condition and inter-camera condition. In~\cite{2018Camera, zhong2019invariance} camera labels are adopted to generate new instances with GAN-based techniques. With contrastive learning, CAP~\cite{Wang2021camawareproxies} adopt camera information during network training, where the memory is divided in proxy level to learn from contrastive pairs with and without the same camera view. However, the pseudo labels of CAP are generated by camera-agnostic clustering. Although various attempts have been made, applying camera variances to generate more reliable pseudo labels under the unsupervised setting is still under-explored. Unlike previous methods, our work aligns the data for each intra-camera and inter-camera condition when computing similarities during clustering, through which we can gain more precise pseudo labels.

\textbf{Contrastive learning.} As a dictionary look-up task, contrastive learning~\cite{wu2018unsupervised,zhuang2019local,he2020momentum} aims to discriminate each instance via treating them as distinct classes. More specifically, an instance-level contrastive loss is adopted to make the feature of an instance close to itself (or augmented feature), and pushed away from other instance features. Although the instance-level contrastive loss can generalize well to downstream tasks, it does not perform well when applied to person re-ID directly. Considering the need to handle intra-class affinities, SPCL~\cite{ge2020selfpaced} adopts contrastive loss for person re-ID and gains impressive results.

\section{Clustering-based Re-ID Baseline}
\label{sec:baseline}
In the context of fully unsupervised person re-ID, we are given a dataset with $N$ person images without any annotations, where $\bX=\left\{x_i\right\}_{i=1}^{N}$. The goal is to train a model that can discriminate identities and generalize well. Our work follows a widely applied clustering-based re-ID pipeline, which iterates between network training with an instance-level memory (an external memory bank) for contrastive learning and clustering to generate pseudo labels. The encoder is initialized by an ImageNet pre-trained model. The encoded features of all the instances via this model are used to construct the instance-level memory. 
After the initialization, the framework alternates between two steps:

(1) Clustering. At the beginning of each epoch, features in the instance-level memory are adopted to compute the distance (cosine similarities) matrix for clustering. Clustering algorithms such as K-means \cite{macqueen1967some}, DBSCAN \cite{ester1996density} 
 \begin{figure}[ht]
   \centering
    \includegraphics[width=\linewidth]{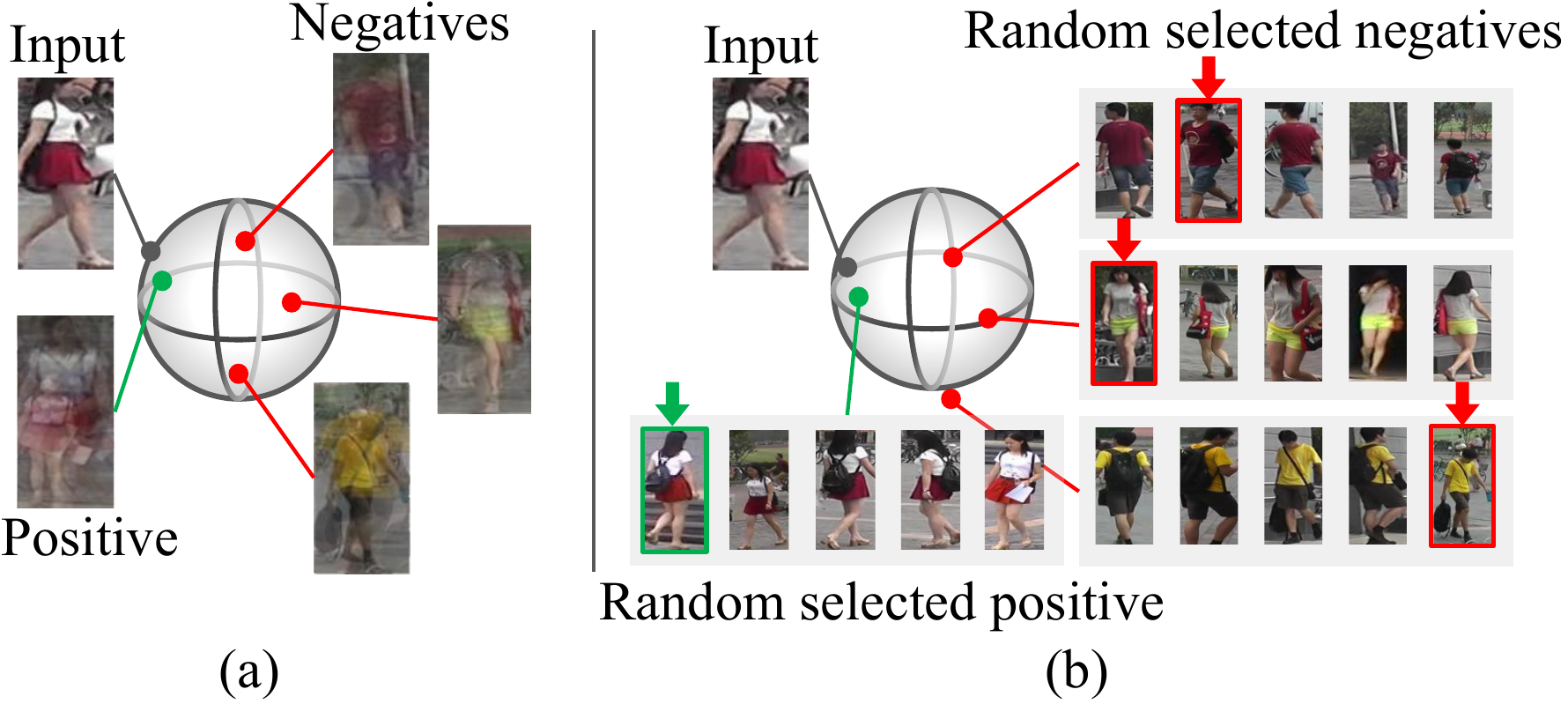}
   \caption{The positive and negative pairs for contrastive learning. (a) depicts the baseline method that adopts mean features of the clusters as the classifier. (b) depicts the stochastic learning strategy that the contrastive pairs are randomly selected among the cluster pools.}
   \label{fig:difference}
\end{figure}
are applied to generate clusters. To generate more accurate pseudo labels, some unreliable images are regarded as outliers that do not belong to any cluster. Specifically, we adopt DBSCAN for clustering, which has two parameters, namely the distance threshold between an instance pair $eps$ and the minimum instance number to form a cluster $min\_num$. For each instance, if there are more than $min\_num$ neighbors whose distance from the instance is less than $eps$, the instance and these neighbors will be assigned in the same cluster. Otherwise, the instance will be an isolated instance. Then, instances that belong to a cluster are assigned pseudo labels according to the clustering result. Following SPCL~\cite{ge2020selfpaced} and CAP~\cite{Wang2021camawareproxies}, we take the isolated instances as outliers, which are regarded as unreliable instances and won't be used for training for this epoch. 
Through this step, the unlabeled dataset with pseudo labels are generated. The dataset is denoted as $\left\{x_i, y_i\right\}_{i=1}^{\hat{N}}$, where $y_i\in\left\{1,...,Y_t\right\}$. $Y$ denotes the number of clusters and is dynamically changed during iterations.


(2) Network training. For each batch, the CNN encoder $f_\theta$ is optimized with pseudo labeled dataset and a contrastive loss:
\begin{small}
\begin{equation}
\begin{aligned}
    \mathscr{L}=-\sum_{i=1}^{N}\log \frac{ \exp{(\langle f_\theta(x_i), c[y_i]\rangle/\tau)}}{\sum_{j=1}^{Y} \exp{(\langle f_\theta(x_i), c[j]\rangle/\tau)} },
\label{loss}
\end{aligned}
\end{equation}
\end{small}

\noindent where $c[j]$ is the cluster centroid corresponding to pseudo label $j$, which is computed as the mean vector of memory features $w$ that belong to this cluster. $\bm{w} \in \mathbb{R}^{N \times d}$ stores the feature of each instances. $\tau$ is a temperature factor, and $\langle\ ,\ \rangle$ indicates the inner product of two feature vectors to measure their similarity. The contrastive loss makes an instance close to the centroid of its cluster, and pushed away from other clusters.

After the backward procedure of each batch, the memory is updated with the encoded features:
\begin{equation}
    w_i=\mu \cdot w_i+(1-\mu)f_{\theta}(x_i),
\end{equation}

\noindent where $i$ means the instance index of $x$ in the memory, and $\mu$ is a momentum factor.

\section{Method}
In this section, we introduce our method in detail. Following the pipeline discussed in Section \ref{sec:baseline}, our framework contains two procedures, \textit{i.e.} network training with pseudo labels and clustering. In this paper, we re-think the elements that affect contrastive learning for unsupervised re-ID and perform network training with a stochastic learning strategy. As for clustering, two efforts are made to boost its performance: (1) Temporal ensembling is adopted to improve the quality of feature embeddings for clustering; (2) A unified distance matrix is proposed to reduce the camera variance while emphasizing the identity variance. The overview of our framework is illustrated in Figure \ref{fig:overview}. 

\subsection{Stochastic Learning Strategy}
As shown in Fig.~\ref{fig:difference}~(a), in the baseline approach, during training, the average feature of each cluster is used as positive and negatives for contrastive loss. However, the pseudo label generated by clustering could be noisy, the negative samples are monotonous and the features in memory are inconsistent due to the change of encoder. To alleviate these defects, We propose a stochastic learning strategy. As shown in Fig.~\ref{fig:difference}~(b), for each training sample, we use last seen instances from each cluster as the positive and negatives. In this way, the effect of clustering error is reduced, more negative samples are observed by the network for more robust training, and the features can be updated in time.

Specifically, we adopt a stochastically-updated memory, which is updated with random instance features for each cluster along with network training. The features of the stochastic memory are used as the classifier for contrastive loss. At the beginning of each iteration, we construct a memory bank $\bm{M} \in R^{Y\times d}$ to represent each cluster, where $Y$ is the cluster number, $d$ is the feature dimension, and $m_i$ denotes the feature of the $i$-th class. 

\textbf{Initialization.} To construct the positive and negative samples for each training sample, the stochastically-updated memory $\bm{M} \in R^{Y\times d}$ is maintained to represent each cluster. 

\textbf{Updating.} For each $P \times K$ batch, we sample $P$ identities and $K$ instances for each person identity. We adopt the sampler in \cite{ge2020selfpaced}, which guarantees that images for each identity are taken by at least two different cameras.

During training, 
the instance $x_i$ is used to update the corresponding cluster center $y_i$. The stochastic memory is updated with encoded feature extracted by the current network:
\begin{equation}
    m_{y_i} = \mu_s \cdot m_{y_i} + (1-\mu_s)f_{\theta}(x_i),
\label{stochastic_update}
\end{equation}
where $\mu_s$ is a momentum factor.


\textbf{Contrastive loss.} The formation of the contrastive loss is the same as Equation \ref{loss}:
\begin{small}
\begin{equation}
\begin{aligned}
    \mathscr{L}=-\sum_{i=1}^{N}\log \frac{ \exp{(\langle f_\theta(x_i), m_{y_i}\rangle/\tau)}}{\sum_{j=1}^{Y} \exp{(\langle f_\theta(x_i), m_j\rangle/\tau)} } ,
\end{aligned}
\end{equation}
\end{small}


Through initializing and updating, a stochastic memory is maintained that represents each cluster with the last seen images. By adopting the stochastic memory, the positive and negative samples are expressed by diverse random instances and are changed along with network training. With the contrastive loss, for each training sample, the relationship between the randomly built positive and negative pairs is learned. Besides, representing positive and negative samples with the last seen instances for each cluster can bring better consistency for contrastive loss.

\subsection{Temporal Ensembling based Camera-aware Clustering}
\textbf{Temporal ensembling based feature embedding.}
As the training targets are obtained by clustering, during training, some positive and negative samples can be expected to be noisy. Then the noise for the final network and its extracted features is inevitable. To build more robust feature for clustering, temporal ensembling \cite{laine2016temporal} is adopted to aggregate the features of multiple previous networks into an ensemble feature embedding.

To retain temporal features for each embedding, we maintain an instance-level memory to represent instance features. Although stochastically-updated memory performs better as a classifier, it loses instance history information among epochs because the cluster changes every epoch. We combine two-level memories. Instance-level memory maintains an exponential moving average (EMA) feature for each of the training samples. During training, the EMA features are updated with up-to-date features. Consequently, the EMA feature of each instance is formed by an ensemble of the feature extracted from the model's current version and those earlier versions. This ensembling improves the quality of the features, and the features will be further used to generate pseudo labels and re-initialize the stochastic memory at the beginning of each iteration.

We store the temporal features $\bm{V} \in \mathbb{R}^{N \times d}$, where $v_i$ is the $i$-th row of $\bm{V}$ indicating the feature of $i$-th instance. The instance-level memory is initialized only once during the training stage by the ImageNet~\cite{2012ImageNet} pre-trained model. At the beginning of each epoch, features in instance memory will be used to compute the distance matrix, which will be clustered to generate pseudo labels.

For each training instance $x_i$, the temporal ensembling based feature is updated with the newly encoded feature $f_{\theta}(x)$:
\begin{equation}
    v_i = \mu_t \cdot v_i + (1-\mu_t)f_{\theta}(x),
\label{temporal_update}
\end{equation}
where $\mu_t$ is a momentum factor.

Because outliers are discarded, their corresponding features won't be updated during this epoch. At the end of epoch, we update outlier features by encoder to retain the consistency of the temporal features.

\textbf{Unified distance matrix for camera aware clustering.}
At each iteration, we compute the distance matrix with temporal ensembling based features, then apply the clustering algorithm to group target instances into clusters and un-clustered instances. Since images across different cameras are subject to different distribution, for some sample pairs, the inter-camera variance can be more severe than the inter-identity variance. Instances may be assigned to the wrong clusters because of camera variances, which will do harm to the model training. When computing the distance matrix and generating pseudo labels, it is essential to take camera variances into consideration. 

In this paper, we propose a unified distance matrix to conduct camera-aware clustering. We denote the corresponding camera ID of each training sample as $\left\{c_i\right\}_{i=1}^{N}$. To efficiently alleviate the influence of the camera variance, we set a camera domain offset matrix $D^{cam}\in N_{cam}\times N_{cam}$ reflects the image differences between different camera domains (or in a domain), where $N_{cam}$ denotes the number of cameras. At each iteration, the matrix $D^{cam}$ is computed by the mean similarity of instance pairs belong to the corresponding camera domain. Then a unified distance $\tilde{D}$ is computed by the original similarity matrix $D\in N\times N$ and the offset matrix $D^{cam}$, where we try to reduce the camera variance $D^{cam}$ from $D$. 
Then in $\tilde{D}$ the distance between instances would better reflect the variance between identities. 

\begin{algorithm}[ht]
\caption{Framework of our method.}
\label{alg:Framework}
\begin{algorithmic}[1]
\Require
Unlabeled dataset $\bm{X}=\left\{x_i\right\}_{i=1}^{N}$, corresponding camera labels $\left\{c_i\right\}_{i=1}^{N}$, CNN encoder $f_{\theta}$
\Ensure
Trained encoder $f_{\theta}$;
\State Initialize the instance-level memory $\bm{V}$ with $f_{\theta}$;
\For{$iter = 1$ to $num\_iters$}
\State Compute the similarity distance $D$ by $\bm{V}$;
\State Compute the Unified distance matrix $\hat{D}$ with camera labels  $\left\{c_i\right\}_{i=1}^{N}$ and $D$;
\State Perform clustering on $\hat{D}$ and remove outliers to generate pseudo labeled dataset  $\left\{x_i, y_i\right\}_{i=1}^{N}$;
\State Initialize the stochastic memory $\bm{M}$ with $\bm{V}$;
\For{$b =1$ to $num\_batches$}
\State Sample mini-batch images;
\State Extract the features of the samples;
\State Compute the loss in Eq.(\ref{loss});
\State Backward to update encoder $f_{\theta}$;
\State Update corresponding cluster center $y_i$ in $\bm{M}$ ;
\State Update corresponding instance feature $x_i$ in $\bm{V}$ ;
\EndFor
\EndFor
\end{algorithmic}
\end{algorithm}

The original distance matrix $D\in N\times N$ is calculated by the cosine similarities between the temporal ensembling based features $\bm{V}$, where $D_{i,j}$ denotes the similarities between the $i_{th}$ instance and the $j_{th}$ instance. 
While the number of identities is far greater than the number of camera domains, we can expect that the mean value of similarities that belong to each camera pairs implicitly denotes the variances caused by corresponding inter-domain or intra-domain conditions (or similarities affected by camera labels). Then the camera variance between camera $i$ and camera $j$ in the camera domain offset matrix is calculated by:
\begin{equation}
    D^{cam}_{i,j}= \frac{1}{N_{cam}^i\times N_{cam}^j} \sum_{i=1}^{N_{cam}^i} \sum_{j=1}^{N_{cam}^j} D_{\left\{u,v|c_u=i,c_v=j\right\}},
\end{equation}

\noindent where $c_u$ denotes the camera label of the $u_{th}$ instance, $N_{cam}^i$ denotes the number of instances captured by camera $i$. Note that when $i$ equals to $j$, the offset distance $D^{cam}_{i,j}$ represents the intra-camera similarity value. A greater value represents more similar domains, and the values of intra-domain (same camera label) should be the largest. We subtract the corresponding value from the distance matrix $D$ as a penalty item. Then the element ${\tilde{D}}_{u,v}$ in $\tilde{D}$ is calculated by:
\begin{equation}
    {\tilde{D}}_{u,v}=D_{u,v}-\lambda \cdot D^{cam}_{c_u,c_v},
\end{equation}
\noindent where $\lambda$ is a factor to control the effect of camera variance. For the dataset with significant camera variance, $\lambda$ will be set larger accordingly.

\begin{table*}[ht]
\centering
\small
\begin{tabular}{l|c||cc||cc||cc||cc}
\hline
\multirow{2}{*}{Methods} &
\multirow{2}{*}{Reference} &
\multicolumn{2}{c||}{Market-1501}      & \multicolumn{2}{c||}{DukeMTMC-reID}      & \multicolumn{2}{c||}{MSMT17}    
       & \multicolumn{2}{c}{VeRi-776}\\
\cline{3-10}  & & mAP & Rank-1 & mAP & Rank-1 & mAP & Rank-1& mAP & Rank-1  \\ \hline
\multicolumn{9}{l}{\textit{Supervised Methods}}    \\\hline
Supervised upper bound & This paper & 85.5 & 94.6  & 77.0 & 88.4  & 52.4 & 77.8 & - & - \\ 
PCB~\cite{sun2018beyond} & ECCV18 & 81.6 & 93.8  & 69.2 & 83.3  & - & - & - & -\\ 
RGA-SC~\cite{zhang2020relation} & CVPR20 & 88.4 & 96.1  & - & -  & 57.5 & 80.3 & - & -\\ \hline

\multicolumn{9}{l}{\textit{Fully Unsupervised}}\\\hline
BUC~\cite{lin2019bottom}  & AAAI19 & 30.6 & 61.0  & 21.9 & 40.2  & - & - & - & -  \\
StyleBUC~\cite{lin2020unsupervised2}& TIP20 & 38.0 & 73.7  & 30.6 & 56.1  & - & - & - & -  \\
UGA~\cite{wu2019unsupervised}  & ICCV19 & 70.3 & 87.2  & 53.3 & 75.0  & 21.7 & 49.5 & - & -  \\
SSL~\cite{lin2020unsupervised} & CVPR20 & 37.8 & 71.7 & 28.6 & 52.5 & - & - & - & -\\
JVTC$^*$~\cite{2020UnJoint} & ECCV20 & 41.8 & 72.9  & 42.2 & 67.6  & 15.1 & 39.0 & - & -\\
MMCL$^*$~\cite{2020Unsupervised} & CVPR20 & 45.5 & 80.3 & 40.2 & 65.2 & 11.2 & 35.4 & - & -     \\
HCT~\cite{zeng2020hierarchical} & CVPR20 & 56.4 & 80.0 & 50.7 & 69.6 & - & - & - & -\\
Cycas~\cite{wang2020cycas} & ECCV20 & 64.8 & 84.8 & 60.1 & 77.9 & 26.7 & 50.1 & - & -     \\
SPCL$^*$~\cite{ge2020selfpaced} & NeurIPS20 & 73.1 & 88.1 & - & - & 19.1 & 42.3 & 36.9 & 79.9 \\
CAP~\cite{Wang2021camawareproxies} & AAAI21 & 79.2 & 91.4 & 67.3 & 81.1 & 36.9 & 67.4 & - & - \\
RLCC~\cite{zhang2021refining} & CVPR21 & 77.7 & 90.8 & 69.2 & 83.2 & 27.9 & 56.5 & 39.6 & 83.4 \\
ICE~\cite{Chen_2021_ICCV} & ICCV21 & 82.3 & \textbf{93.8} & 69.9 & 83.3 & 38.9 & 70.2 & - & - \\
\hline
Baseline& This paper   & 73.8 & 88.9 & 63.5 & 78.8 & 21.1 & 46.0 & 33.1 & 74.9   \\
Ours w$\backslash$ ResNet50  & This paper & \textbf{82.4} & 93.0  & 72.2 & \textbf{84.9} & 38.4 & 68.6 & 43.2 & 87.0 \\
Ours w$\backslash$ ResNet50-IBN  & This paper & 82.0 & 92.8 & 72.7 & 84.8 & \textbf{42.4} & \textbf{71.6} & \textbf{43.9} & \textbf{88.9} \\
\hline

\end{tabular}
\vspace{0.1cm}
\caption{Comparison with state-of-the-art unsupervised re-ID methods on four datasets. SOTA fully unsupervised and supervised methods are shown. The best results among unsupervised methods are marked in bold type. $^*$ denotes the UDA-based method working under the fully unsupervised setting.}
\label{tab:SOTA}
\end{table*}

After getting the new distance matrix $\tilde{D}\in N\times N$, softmax is applied to each line of the matrix. In this way, the distance between instances is represented as the relative distance. Then the unified distance matrix $\tilde{D}$ is adopted for clustering algorithm to generate pseudo labels. By taking each intra-camera and inter-camera pair into consideration, we can align instances in the distance space and generate pseudo labels with a new distance matrix that is more related to the variances of identities.

\subsection{Summary of the framework}
The framework iteratively trains the network and conducts clustering to generate pseudo labels for further network updating. In this framework, we consider the factors that improve contrastive learning and clustering. During training, the network is learned from random and up-to-date positive and negatives, which reduces the effect of label noise from clustering and brings better consistency. For clustering, the unified distance matrix is proposed to deal with the camera variance and temporal ensembling is adopted to provide a more robust feature for clustering. 
The whole procedure is summarized in Algorithm \ref{alg:Framework}.

\section{Experiments}

\begin{table*}[ht]
\centering
\small
\begin{tabular}{l|c||ccc||ccc}
\hline
\multirow{2}{*}{Methods} &
\multirow{2}{*}{Reference} &
\multicolumn{3}{c||}{DukeMTMC-reID → Market-1501}      & \multicolumn{3}{c}{Market-1501 → DukeMTMC-reID}          \\ \cline{3-8}  & & mAP & Rank-1& Rank-5 & mAP & Rank-1& Rank-5  \\ \hline
ECN~\cite{zhong2019invariance} & CVPR19 & 43.0 & 75.1 & 87.6 & 40.4 & 63.3 & 75.8 \\ 
SSG~\cite{fu2019self} & ICCV19 & 58.3 & 80.0 & 90.0 & 53.4 & 73.0 & 80.6  \\
JVTC~\cite{2020UnJoint} & ECCV20 & 61.1 & 83.8 & 93.0 & 56.2 & 75.0 & 85.1\\
DG-Net++~\cite{2020Joint} & ECCV20 & 61.7 & 82.1 & 90.2 & 63.8 & 78.9 & 87.8 \\
ECN+~\cite{2020Learning} & PAMI20 & 63.8 & 84.1 & 92.8 & 54.4 & 74.0 & 83.7 \\
MMT~\cite{ge2020mutual} & ICLR20 & 71.2 & 87.7 & 94.9 & 65.1 & 78.0 & 88.8 \\
SPCL~\cite{ge2020selfpaced} & NeurIPS20 & 76.7 & 90.3 & 96.2 & 68.8 & 82.9 & 90.1\\
Dual-Refinement~\cite{dai2021dual} & TIP21 & 78.0 & 90.9 & 96.4 & 67.7 & 82.1 & 90.1 \\
GLT~\cite{zheng2021group} & CVPR21 & 79.5 & 92.2 & 96.5 & 69.2 & 82.0 & 90.2 \\
IDM~\cite{dai2021idm} & ICCV21 & 82.8 & 93.2 & 97.5 & 70.5 & 83.6 & 91.5 \\ \hline
Ours (unsupervised)& This paper & 82.4 & 93.0 & \textbf{97.5} & 72.2 & \textbf{84.9} & \textbf{92.3}
\\ 
Ours (UDA)& This paper & \textbf{84.0} & \textbf{93.7} & 97.4 & \textbf{72.6} & 84.7 & 91.9
\\ \hline
\\ \hline

\multirow{2}{*}{Methods} &
\multirow{2}{*}{Reference} &
\multicolumn{3}{c||}{Market1501 → MSMT17}      & \multicolumn{3}{c}{DukeMTMC-reID → MSMT17}          \\ \cline{3-8}  & & {\quad}mAP & {\quad}Rank-1& Rank-5 & {\quad}mAP & {\quad}Rank-1& Rank-5  \\ \hline
ECN~\cite{zhong2019invariance} & CVPR19 & 8.5 & 25.3 & 36.3 & 10.2 & 30.2 & 41.5 \\
SSG~\cite{fu2019self} & ICCV19 & 13.2 & 31.6 & - & 13.3 & 32.2 & -  \\
JVTC~\cite{2020UnJoint} & ECCV20 & 19.0 & 42.1 & 53.4 & 20.3 & 45.4 & 58.4\\
DG-Net++~\cite{2020Joint} & ECCV20 & 22.1 & 48.4 & 60.9 & 22.1 & 48.8 & 60.9 \\
ECN+~\cite{2020Learning} & PAMI20 & 15.2 & 40.4 & 53.1 & 16.0 & 42.5 & 55.9 \\
MMT~\cite{ge2020mutual} & ICLR20 & 22.9 & 49.2 & 63.1 & 23.5 & 50.5 & 63.6 \\
SPCL~\cite{ge2020selfpaced} & NeurIPS20 & 25.4 & 51.6 & 64.3 & 26.5 & 53.1 & 65.8\\
Dual-Refinement~\cite{dai2021dual} & TIP21 & 25.1 & 53.3 & 66.1 & 26.9 & 55.0 & 68.4 \\
GLT~\cite{zheng2021group} & CVPR21 & 26.5 & 56.6 & 67.5 & 27.7 & 59.5 & 70.1 \\
IDM~\cite{dai2021idm} & ICCV21 & 33.5 & 61.3 & 73.9 & 35.4 & 63.6 & 75.5 \\ \hline
Ours (unsupervised) & This paper & 38.4 & 68.6 & 79.4 & 38.4 & 68.6 & 79.4\\
Ours (UDA) & This paper & \textbf{40.9} & \textbf{70.3} & \textbf{80.9} & \textbf{42.4} & \textbf{71.7} & \textbf{81.7}
\\ \hline
\end{tabular}
\caption{Comparison with state-of-the-art UDA methods on three datasets. The best results among unsupervised domain adaptive methods are marked in bold.}
\label{tab:UDA}
\end{table*}

\subsection{Datasets and Evaluation Protocol}
We evaluate the proposed method on three large-scale person re-ID benchmarks, including Market-1501, DukeMTMC-reID and MSMT17 and a vehicle re-ID dataset VeRi-776.

\textbf{Market-1501}~\cite{zheng2015scalable} includes 32,668 labeled person images of 1,501 identities collected from six camera views. For evaluation, the dataset is divided into 12,936 images of 751 identities for training, 3,368 query images and 19,732 images of 705 identities for testing.

\textbf{DukeMTMC-reID} is a subset of the DukeMTMC dataset~\cite{ristani2016performance}. The dataset is captured from eight cameras, including 36,411 person images from 1,812 identities. A number of 1,404 identities appear in more than two cameras, and the rest 408 IDs are distractor images. Using the evaluation protocol specified in \cite{zheng2017unlabeled}, the dataset is divided with 16,522 images of 702 identities for training, 2,228 query images of 702 identities and 17,611 gallery images for testing.

\textbf{MSMT17}~\cite{wei2018person} is composed of 126,411 person images from 4,101 identities collected by an 15-camera system. These cameras include 12 outdoor cameras and 3 indoor ones. The training set consists of 32,621 images of 1,041 identities, and the testing set contains 11,659 images as query and 82,161 images as gallery. With large-scale images and multiple cameras, this dataset is more challenging than the other two benchmarks.

\textbf{VeRi-776}~\cite{liu2016deep} collects vehicle images in the real-world urban surveillance scenario. The training set has 575 vehicles with 37, 746 images and the testing set has 200 vehicles with 11, 579 images, captured by 20 cameras.

\textbf{Evaluation protocol.} Mean average precision (mAP), cumulative matching characteristic (CMC) and the clustering accuracy are adopted to measure the performance. The presumption is that CMC reflects retrieval precision, while MAP reflects the recall. The clustering accuracy is computed as the average proportion of the images of the majority identity in each cluster. No post processing technique like re-ranking is used.

\textbf{Implementation Details.} We adopt an ImageNet-pretrained ResNet-50~\cite{he2016deep} as the backbone of encoder $f_\theta$. The $L2$ normalized feature is used to update instance-level memory and cluster-level memory. The update rate is empirically set as $0.2$, the temperature factor is $0.04$. At each iteration, we adopt DBSCAN to cluster features from instance memory and generate pseudo labels. DBSCAN has two important parameters: distance threshold between an instance pair and minimum instance number to form a cluster. We set the minimum number as 4 for all three datasets, the threshold of DBSCAN is $0.5$ for DukeMTMC-reID and Market-1501, $0.7$ for MSMT17 and VeRi-776.

We use ADAM as the optimizer and train the model for 80 epochs. The initial learning rate is 0.00035. Each training batch consists of 64 images randomly sampled from 16 identities with 4 images per identity. Random flipping, cropping and erasing are applied as data augmentation.

\subsection{Comparison with State-of-the-Arts.}
In this subsection, we compare our methods with the state-of-the-arts unsupervised and UDA re-ID methods. The results are shown in Table \ref{tab:SOTA} and Table \ref{tab:UDA} respectively. We also report our performance with ResNet50-IBN, which is formed by replacing all BN layers in ResNet-50~\cite{he2016deep} with IBN~\cite{pan2018two} (Instance-batch normalization) layers.

\begin{table*}[ht]
\centering
\small
\begin{tabular}{l||ccc||ccc||ccc}
\hline
\multirow{2}{*}{Methods} & \multicolumn{3}{c||}{Market-1501}      & \multicolumn{3}{c||}{DukeMTMCreID}      & \multicolumn{3}{c}{MSMT17}    \\ \cline{2-10} 
                         & mAP & Rank-1& Rank-5 & mAP & Rank-1& Rank-5 & mAP & Rank-1& Rank-5  \\ \hline
Baseline  & 73.8 & 88.9 & 94.9 & 63.5 & 78.8 & 87.8  & 21.1 & 46.0 & 57.7   \\

Stochastic & 80.3 & 92.0 & 96.9 & 69.5 & 83.5 & 90.8 & 30.6 & 59.8 & 72.5\\
Stochastic (online) & 81.5 & 92.5 & 96.7 & 70.8 & 84.5 & 91.8 & 32.6 & 62.8 & 74.4\\
Hard & 80.6 &91.6& 96.7 &70.1 & 82.6 & 90.6 & 30.8 & 59.1 & 71.2 \\
+Temporal & 82.2 & 92.9 & 97.1 & 71.2 & 84.3 & 92.1 & 33.6 & 63.2 & 74.8      \\
Hard+Temporal\&Cam & 82.1 & 92.3 & 96.9 & 71.8 & 84.5 & 91.9 & 34.7 & 64.0 &75.2\\
+Temporal\&Cam & 82.4 & 93.0 & 97.5 & 72.2 & 84.9 & 92.3 & 38.4 & 68.6 & 79.4 \\
\hline
\end{tabular}
\vspace{0.1cm}
\caption{Ablation studies on three datasets. `Stochastic' stands for using stochastic training strategy. `Stochastic (online)' stands for using stochastic training strategy and update the memory by the up-to-date feature embedding online to maintain the consistency. `Hard' stands for using hard sampling mining to update the cluster based memory. `Temporal' refers to conduct temporal ensembling based feature embedding for clustering. `Cam' stands for alleviating the influence of camera variances for clustering by the unified distance matrix.}
\label{tab:Ablation}
\end{table*}

\textbf{Fully unsupervised person re-identification.} 
Our method has competitive performance on all three datasets. On Market-1501, we obtain 82.4\% of mAP and 93.0\% of rank-1 accuracy, which is close to the supervised upper bound. 
Compared with the baseline, the proposed method can significantly improve the performances on three datasets. The full model gains 4.1\% Rank-1 and 8.5\% mAP improvements on Market-1501, 6.1\% Rank-1 and 8.7\% mAP improvements on DukeMTMC-reID. Besides, it gets 22.6\% Rank-1 and 17.3\% mAP improvements on the most challenging MSMT17, which has complex identity variances and camera variances. 

Compared with CAP~\cite{Wang2021camawareproxies}, we achieve 3.2, 4.9 and 1.5 points of improvement in terms of mAP on Market-1501, DukeMTMC-reID, and MSMT17, respectively. Compared with ICE~\cite{Chen_2021_ICCV}, we achieve comparable performance on Market-1501 and DukeMTMC-reID, and is 0.5\% point lower on MSMT17. In CAP~\cite{Wang2021camawareproxies} and ICE~\cite{Chen_2021_ICCV}, proxy-level memory bank is adopted to conduct both the inter and intra-camera contrastive learning. Compared with these method, our stochastic learning strategy is more straight forward and the camera-aware clustering can be view as a flexible add-on for future works. 

\begin{figure*}[ht]
   \centering
   \includegraphics[width=\linewidth
   ]{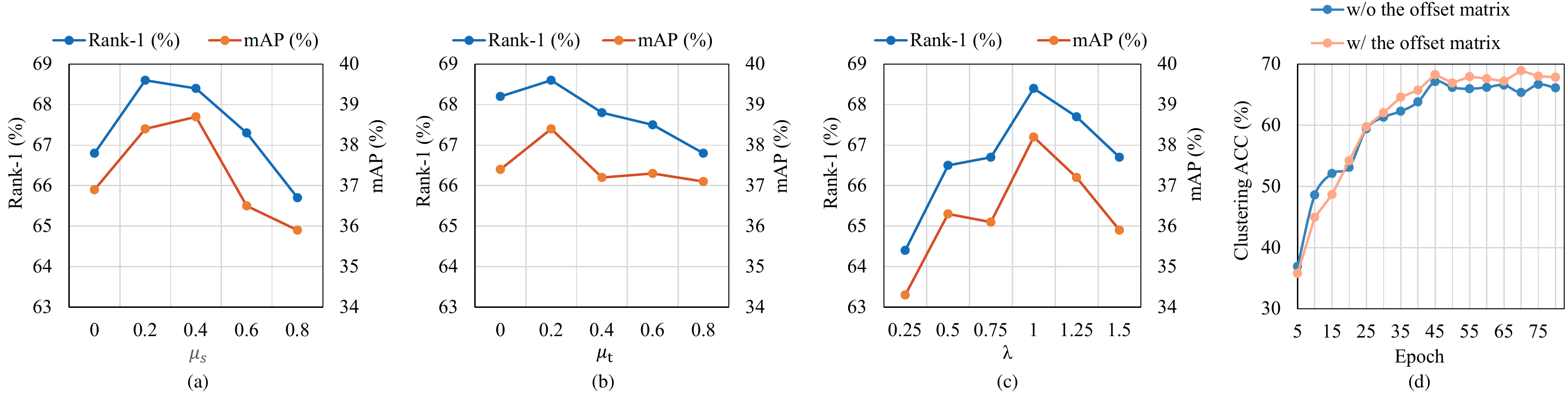}
   \caption{Algorithm analysis on MSMT17. (a) Evaluations of the momentum update factor $\mu_s$. (b) Evaluations of the momentum update factor $\mu_t$. (c) Evaluations of the camera offset factor $\lambda$. (d) Clustering accuracy along training on MSMT17. The clustering accuracy with and without camera offset is shown.}
   \label{fig:hyperpara}
\end{figure*}

\textbf{Fully unsupervised vehicle re-identification.} Since our method is general for object re-ID, we also evaluate our method on VeRi-776 for vehicle re-identification. The results are reported in Table~\ref{tab:SOTA}. Our method outperforms the second place method RLCC~\cite{zhang2021refining} with 4.3\% mAP and 5.5\% Rank-1 gain. The consistent performance improvement on various datasets demonstrates that our method is effective over general object re-identification tasks.

\textbf{UDA-based person re-identification.} For the UDA setting, the target domain is treated the same as the unsupervised version. As for the source domain, we store and update class centers according to their ground truth labels. Both source class centers and target cluster centers are used to compute the contrastive loss. As shown in Table~\uppercase\expandafter{\romannumeral2}, our method outperforms the state-of-the-art UDA methods on all three datasets. On the challenging Market1501 → MSMT17 and DukeMTMC-reID → MSMT17 setting, our method achieves 7.4\%/ 9\% and 7\%/ 8.1\% mAP/Rank-1 gain compared to the SOTA method IDM~\cite{dai2021idm}.

\textbf{Comparison with Supervised methods.} We report our supervised upper bound in Table \ref{tab:SOTA}, in which the model is trained with our backbone and the annotated labels. Comparing with the upper bound, our method achieves competitive performance on Market-1501 and DukeMTMC, while the performance gap is relative large on the most  challenging dataset MSMT17. 
We also observe that, comparing with the classic supervised method  PCB~\cite{sun2018beyond}, our unsupervised method achieves 0.8 and 3.0 points of improvement in terms of mAP on Market-1501 and DukeMTMC-reID, respectively. The remarkable performance demonstrates our superiority. 

\subsection{Ablation Studies}
In this subsection, we analyze the sub-components of our method and show the performances in Table \ref{tab:Ablation}. The hyper-parameters are also evaluated and discussed.

\textbf{Effectiveness of the stochastic learning strategy.}
Three experiments are conducted to show how the stochastic learning strategy benefits the model. (1) Baseline: the mean vector of instance features of each cluster is used as a classifier for contrastive loss. 
(2) Stochastic: Store and update each cluster center as classifiers. For each input instance, a randomly selected feature embedding of the corresponding cluster is used to update the cluster center. (3) Stochastic (online): Store and update each cluster center as classifiers. Each input embedding is used to update its corresponding cluster center, so cluster centers are updated with the last seen instances. The results are shown in Table~\ref{tab:Ablation}.

\begin{figure*}[ht]
   \centering
    \includegraphics[width=0.85\linewidth]{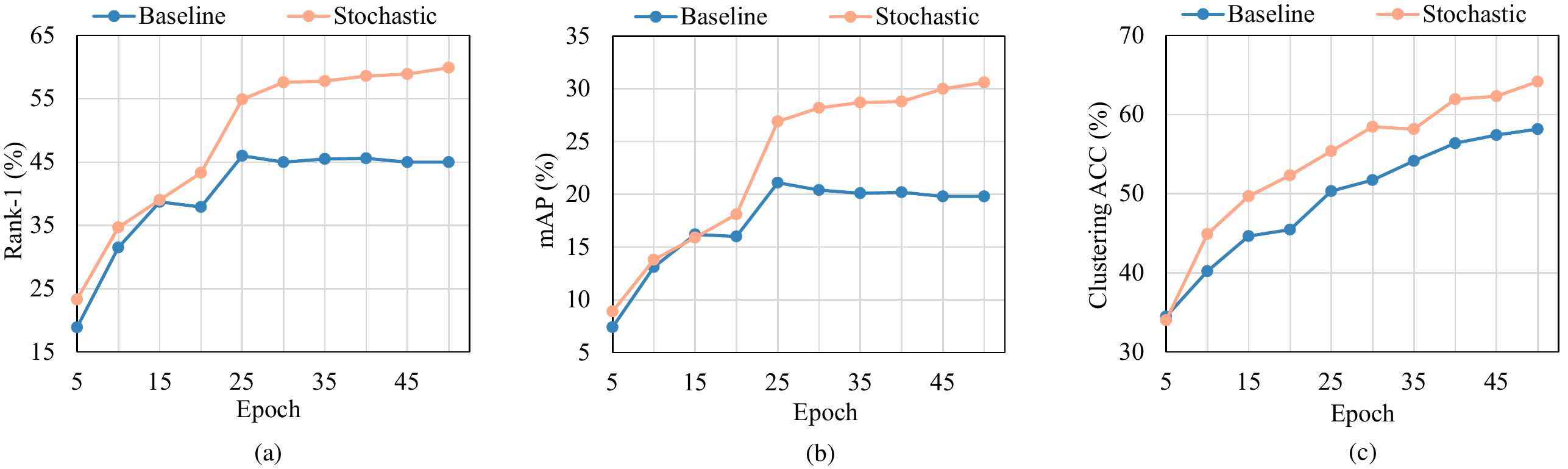}
   \caption{Performances along training on MSMT17. The mAP, rank-1 accuracy and the accuracy of clustering along training are shown in the sub-figure (a), (b) and (c), respectively. The performances of the baseline and the method using stochastic training strategy are shown. Note that `Stochastic' stands for using stochastic training strategy without online feature updating temporal ensembling or unified matrix. }
   \label{fig:exp_iteration}
\end{figure*}

\begin{figure}[ht]
   \centering
   \includegraphics[width=0.7\linewidth
   ]{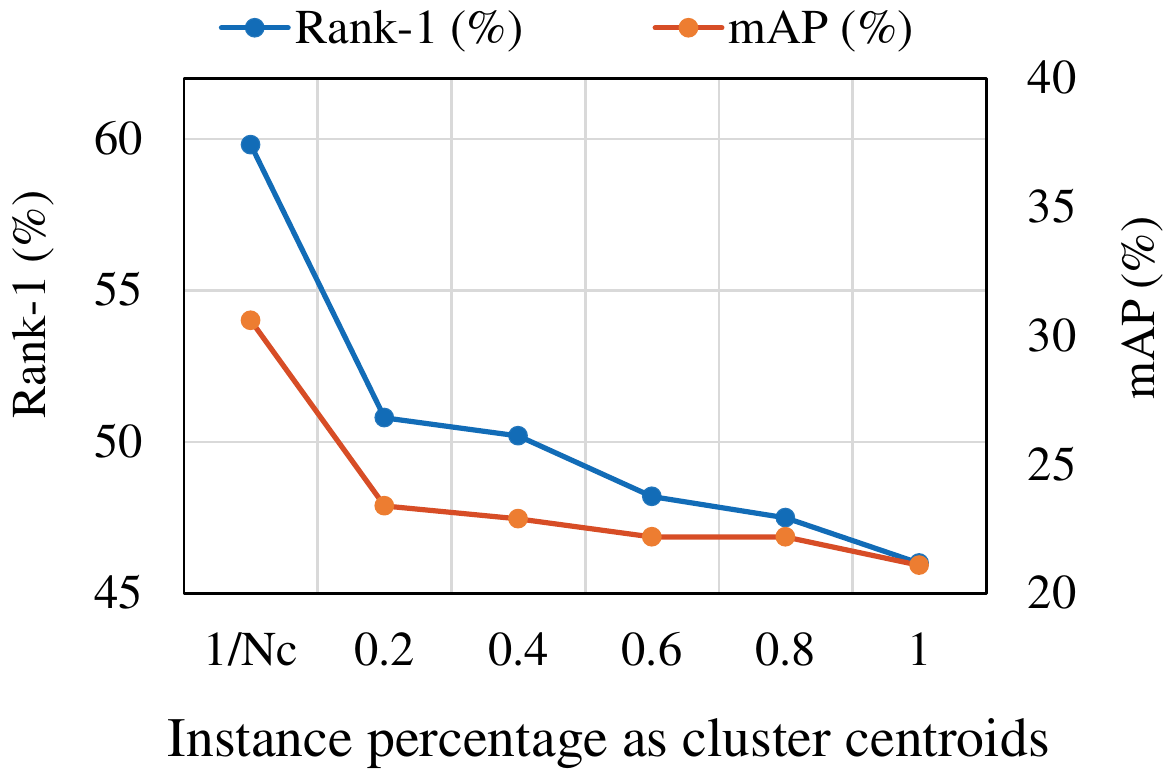}
   \caption{Evaluations with different instance percentages for centroids on MSMT17. The rank-1 accuracy and mAP are shown. $N_c$ denotes the number of images in each cluster.}
   \label{fig:stochastic}
\end{figure}

We observe that the performance of baseline is obviously lower than the other two stochastic methods. The main reason is that due to the imperfection of the encoder and clustering algorithm, the error of pseudo labels is inevitable. Thus, the baseline method could accumulate the clustering errors among the training-clustering iterations. Taken random sample to represent each cluster centroids can greatly alleviate the influence of noisy label. Also, we suspect that, the mean operation harms the diversity of the positive and negative samples when conduct contrastive learning, which would impair the model.

Compared with `Stochastic', `Stochastic (online)' brings further improvements on all the three datasets, \textit{e.g.} 2\% and 3\% performance gain on mAP and rank-1 accuracy in MSMT17. In `Stochastic', the random selected instance features are encoded by encoder of different periods, which may bring inconsistency when comparing with the current training instance feature to compute contrastive loss. We demonstrate that by storing and updating the stochastic memory online, we can always represent each cluster with newly referred instances, which handles inconsistency and noisy labels simultaneously.

We also explore the momentum update procedure. The performances of different values of momentum factor $\mu_s$ for stochastic memory are shown in Fig. \ref{fig:hyperpara} (a). As in Eq. \ref{stochastic_update}, adopting smaller $\mu_s$ means assigning more weight to the last seen instance feature. If we set $\mu_s$ to 0, then each cluster center will be represented completely with a last seen instance. We can observe that the stochastic learning strategy gets optimal performance when $\mu_s$ is 0.2$\sim$0.4, which achieves a balance between attaching importance to last seen instances and combining with previous representations. We set $\mu_s$ as 0.2 for all four datasets.


\textbf{Stochastic sampling vs hard sampling.} We also evaluate our method with hard sample mining, where instances that have the smallest similarities with their centroids are selected to update the cluster memory instead of the stochastic ones. The performance is shown in Table~\ref{tab:Ablation}. It can be seen that proposing the hardest update solely or with the complete framework will both bring performance degradation. With the whole design, the performance of using hard sample mining is slightly worse with 0.3\% and 0.4\% mAP decreases on Market-1501 and DukeMTMC-reID, respectively. However, on MSMT17, the mAP drops by 3.7\%. We suspect that the hard sampling strategy is affected by unreliable samples, especially when the dataset is challenging, e.g., MSMT17. There are more efforts should be made to select both hard and reliable samples.


\textbf{Effectiveness of the temporal ensembling based feature embedding.} 
The performance of with and without temporal ensembling based feature embedding is shown in Table \ref{tab:Ablation}. 
The temporal ensembling based feature embedding provides performance improvements on all the three datasets. Specifically, on MSMT17, the temporal ensembling based feature embedding improves the mAP and rank-1 accuracy by 1.0 points and 0.4 points, respectively.
The improvement demonstrates that, by considering the features extracted by temporal networks, the assembled feature becomes more robust and performs better during clustering. 

The performances of different values of momentum factor $\mu_t$ for instance memory are shown in Fig. \ref{fig:hyperpara} (b). As in Eq. \ref{temporal_update}, a smaller $\mu_t$ means assigning more weight to the last seen instance. It achieves optimal performance when $\mu_t$ is around 0.2. However, when $\mu_t$ continuously gets larger (more history features are considered), the performance gets worse. This is reasonable, because generally more up-to-date feature behaves better, and when the old feature dominates, the performance is affected. We set $\mu_t$ as 0.2 for all four datasets.

\textbf{Effectiveness of unified distance matrix.} Table \ref{tab:Ablation} shows the performance of with and without the unified distance matrix. We observe that takes camera information for distance matrix outperforms the counterpart on all the three datasets. Specifically, the improvement of mAP on the three datasets are 0.2\%, 1.0\%, 4.8\%, respectively. We observe that, the effect is quite significant on MSMT17, which has the most cameras and the most severe camera variances. The improvement demonstrates that the proposed matrix successfully reduces the affect of camera variances, but focuses on the variance of person identities. The component is easily be add-on other clustering based method and works well especially facing challenging dataset.

The clustering accuracy with and without the unified distance matrix along training is shown in Fig. \ref{fig:hyperpara} (d). We observe that at the first 25 epochs, the method without the unified distance matrix achieves a better clustering result. However, when the training epoch gets larger, the method with the unified distance matrix achieves consistent performance gain over the baseline. This demonstrates that when the model or the feature is strong enough, the offset matrix benefits clustering. 

We also evaluate the camera offset factor $\lambda$ in Fig. \ref{fig:hyperpara} (c), which is the weight of camera variances. We vary the value of $\lambda$ to six different values, the mAP and rank-1 accuracy are illustrated. We observe that, too small $\lambda$ may undervalue the weight of camera variances, and too large $\lambda$ will do harm to the computed similarity. When increasing the value, it achieves consistent improvements and gains optimal performance when $\lambda=1$.


\subsection{Algorithm Analysis}
\textbf{Analysis over iterations.} We show the re-ID performance and the clustering accuracy along training in Fig. \ref{fig:exp_iteration}. It is obvious that the stochastic strategy achieves better performance on all of the three criterion. 
We observe that, the rank-1 accuracy and mAP of the two methods are similar in the first 20 epochs, after which the performance gap becomes bigger and consistent. We suspect that in the first 20 epochs, both model are relatively weak and have the room for improvement. However, as the network continues to train, the baseline method is limited by the accuracy of clustering and the diversity of the negative samples. 
We also observe that the clustering accuracy is relatively low, indicating that there is still a large gap to meet towards the supervised upper-bound.

\textbf{Analysis of the stochastic learning strategy.} We further dig out how the stochastic strategy works in contrastive learning in Fig. \ref{fig:stochastic}. In the experiments, we randomly select different percentages of instances from each cluster to calculate their mean features as the positive and negatives during contrastive learning. When the percentage is $1/N_c$, the stochastic learning strategy is adopted, where only one instance is selected. When the percentage is 1, the method converge to the baseline, where the mean feature of all images from each cluster is adopted. We observe that when less instance is involved, a higher performance is obtained, which demonstrates the benefit of the stochastic strategy.

\section{Conclusion.} 
In this paper, we investigate the problem of unsupervised re-ID. Following the clustering-based pipeline, we propose to learn from a stochastic strategy with a stochastic memory so as to avoid training error accumulation from clustering noise, provide more diverse negative samples as well as maintain consistent contrastive samples for the current training image. To cope with camera variance, we introduce a unified distance matrix, which reduces the effect of camera differences and is more related to identity variance. For a stable clustering result, temporal ensembling is adopted to generate more robust feature embedding for clustering. Experiments on four datasets and comparison with state-of-the-art methods confirm the validity of our work.


\noindent {\bf Acknowledgment.} This work was supported in part by the National Natural Science Foundation of China under Grant 62001331, in part by the Natural Science Foundation of Hubei Province under Grant 2021CFB475, in part by the Key Research and Development Program of Hubei Province under Grant 2021BAA187, and in part by the Science and Technology Major Project of Hubei Province (Next-Generation AI Technologies) under Grant 2019AEA170.

{
\bibliographystyle{IEEEtran}
\bibliography{egbib}
}

\begin{IEEEbiography}[{\includegraphics[width=1in,height=1.25in,clip,keepaspectratio]{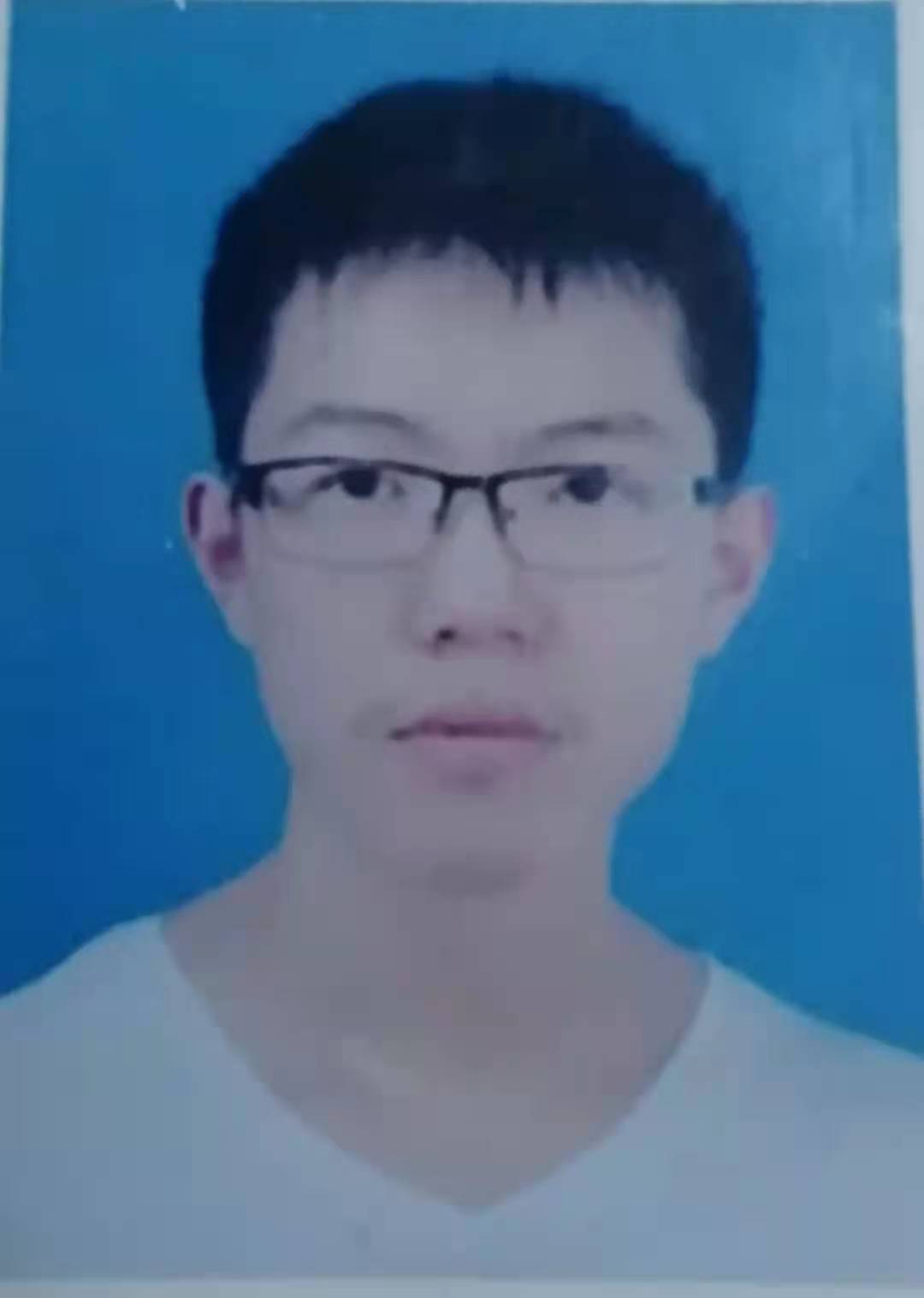}}]{Tianyang Liu} received the B.E. degree from Wuhan university, China, in 2015. He is currently a master student in Wuhan University, China. His research interests are person re-ID and unsupervised learning.    
\end{IEEEbiography}

\begin{IEEEbiography}[{\includegraphics[width=1in,height=1.25in,clip,keepaspectratio]{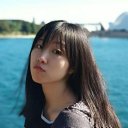}}]{Yutian Lin} received the B.E. degree from Zhejiang University, China, in 2016, and the Ph.D. degree from the Center for Artificial Intelligence, University of Technology Sydney, Australia, in 2019. She is currently an associate professor in Wuhan University, China. Her research interests include person re-ID and related applications, unsupervised learning and self-supervised learning.  
\end{IEEEbiography}

\begin{IEEEbiography}
[{\includegraphics[width=1in,height=1.25in,clip,keepaspectratio]{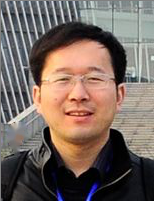}}]{Bo Du} (Senior Member, IEEE) received the Ph.D. degree in photogrammetry and remote sensing from the State Key Lab of Information Engineering in
Surveying, Mapping and Remote Sensing, Wuhan University, Wuhan, China, in 2010. He is currently a Professor with National Engineering Research Center for Multimedia Software, Institute of Artificial Intelligence, Wuhan University, Wuhan, China.His major research interests include pattern recognition, hyperspectral image processing, machine learning, and signal processing.
\end{IEEEbiography}

\end{document}